\documentclass{article} 
\usepackage{nips13submit_e,times}
\usepackage{hyperref}

\usepackage{graphicx}
\usepackage{epsfig}
\usepackage{caption}
\usepackage[labelformat=simple]{subcaption}

\usepackage{cite}
\usepackage{url}
\usepackage{wrapfig}

\nipsfinaltrue  

\begin{document}

\title{An Analysis of the Connections  \\ Between Layers of Deep Neural Networks}

\author{
Eugenio Culurciello\thanks{More information on Eugenio Culurciello's laboratory and research can be found here: http://engineering.purdue.edu/elab/. Real time robotic vision systems: http://www.neuflow.org/} \\
Purdue University\\
\texttt{euge@purdue.edu} \\
\And
Aysegul Dundar \\
Purdue University\\
\texttt{adundar@purdue.edu}
\AND
Jonghoon Jin \\
Purdue University\\
\texttt{jhjin@purdue.edu}
\And
Jordan Bates \\
Purdue University\\
\texttt{jtbates@purdue.edu}
}


\maketitle

\begin{abstract}
We present an analysis of different techniques for selecting the connection between layers of deep neural networks.
Traditional deep neural networks use random connection tables between layers to keep the number of connections small and tune to different image features.
This kind of connection performs adequately in supervised deep networks because their values are refined during the training.
On the other hand, in unsupervised learning, one cannot rely on back-propagation techniques to learn the connections between layers.
In this work, we tested four different techniques for connecting the first layer of the network to the second layer on the CIFAR and SVHN datasets and showed that the accuracy can be improved up to $3\%$ depending on the technique used.
We also showed that learning the connections based on the co-occurrences of the features does not confer an advantage over a random connection table in small networks.
This work is helpful to improve the efficiency of connections between the layers of unsupervised deep neural networks.
\end{abstract}

\section{Introduction}

Most scientists and engineers are fascinated by the design of an artificial vision system that can reproduce some of the human visual system's capabilities of detecting, categorizing, tracking objects in view. 
The availability of a real-time synthetic vision system with such capabilities would find use in a large 
variety of applications, such as: autonomous cars, co-robots helpers, smart appliances, cellular-phones, to name a few. 
The most promising approach in recent years is the fusion of bio-inspired and neuromorphic vision models with machine learning 
\cite{lecun_gradient-based_1998,hadsell_dimensionality_2006,gregor_structured_2011,riesenhuber_hierarchical_1999,serre_feedforward_2007,serre_neuromorphic_2010,jarrett_what_2009,lecun_convolutional_2010,boureau_theoretical_2010}.
This field, named Deep Learning, has provided state-of-the-art results in the categorization 
of multiple objects in static frames \cite{krizhevsky_imagenet_2012}.

Deep Learning networks are computer-vision and computational-neuroscience models of the mammalian visual 
system implemented in deep neural networks, where each network layer is composed of: linear two-dimensional filtering, 
non-linearity, pooling of data, output data normalization \cite{jarrett_what_2009,lecun_convolutional_2010,boureau_theoretical_2010}. 
Deep networks are trained with the abundant image and video content available on the internet and with large labeled datasets. 
In particular, deep networks need to learn good feature representations for complex visual tasks such as object categorization and 
tracking of objects in space and time, identifying object presence and absence.
These representations usually involve learning the linear filter weight values from labeled and unlabeled input data.
Since labeled data is costly and imperfect due to human error 
\cite{karpathy_lessons_2011, torralba_unbiased_2011, hou_meta-theory_2012}, the recent focus is on learning these features purely 
from unlabeled input data \cite{olshausen_emergence_1996, hyvarinen_independent_2000, hinton_fast_2006, vincent_extracting_2008, coates_analysis_2011}.
These recent methods typically learn multiple layers of deep networks by training several layers of features, one layer at a time, with varying complexity of learning models.
Traditional work in unsupervised deep learning has focused on learning layer features, and rarely on the connections between layers.
In this paper, we present an analysis of different connection topologies between layers of deep neural networks for general-purpose vision systems. 

Recent techniques based on unsupervised Clustering Learning (CL) are especially promising because they use 
simple learning methods that quickly converge \cite{culurciello2013clustering,coates_analysis_2011}.
These algorithms are easy to setup and train and are especially suited for applied research because environment-specific data 
can be collected quickly with a few minutes of video and the network can be adapted to specific tasks in minutes.

The paper is organized in the following way: 
Use of clustering learning for learning filters with different connections of deep networks will be explained in the Section \ref{sec-main}. 
The details of the networks including the pre-processing and similarity calculation for learning groups of connections will be described in Section \ref{sec-methods}. 
Finally, we will discuss and analyze our results from CIFAR-10 and SVHN datasets in the Section \ref{sec-results}.

\section{Main idea and contribution}
\label{sec-main}

The connections between layers in a deep neural network are very important parameters. These connections are also referred as \textit{receptive fields} (RF).
It has been shown that in many cases random filters perform only slightly worse than fully trained network layers \cite{saxe2011random} suggesting
that filters, used in the convolutional module of each layer, might not play a dominant role in determining the network performance.
Rather, the connections between layers and the combination of features are mainly responsible for the recent success of 
supervised deep convolutional networks \cite{krizhevsky_imagenet_2012}.
In these networks the connection between layers is
learned from the data using global gradient-descent techniques at large scale.
Because training will intimately select the strength of connections, 
developers of deep networks have traditionally used fully connected layers, custom table connections \cite{lecun_gradient-based_1998}, or 
random connections \cite{lecun_convolutional_2010}.
The reason for using custom RF is to reduce the number of computations, and avoid over-fitting problems.
 
In unsupervised deep networks, one cannot rely on back-propagation techniques to learn the connections between layers. 
This is because unsupervised networks do not used labeled data, and thus cannot compute an error signal. 
Fully connected layers can be used \cite{culurciello2013clustering,coates_analysis_2011} at the expense of more computational time and also reduced performance.
The lower performance is attributable to the averaging effect that a fully connected layer has in an unsupervised network, 
where the connection weights are not adapted to the training data.
In order to use unlabeled data to train large deep networks, we investigate different techniques for grouping features from one layer into the features of the next layers, 
and at the same time learn the filters needed by the next layer.
Figure \ref{fig-learnlayers} qualitatively explains the main contribution of the paper.

In Figure \ref{fig-learnlayers} we show two layers of a deep network: layer $N$ and layer $N+1$.
The maps on the left of the figure are the $N1$ output feature maps of the $N$-the layer in response to an input image.
The maps on the right of the figure are the $N2$ outputs of the first convolutional module of layer $N+1$. 

\subsection{Grouping features into receptive fields}

\begin{figure}
\includegraphics[width=5in]{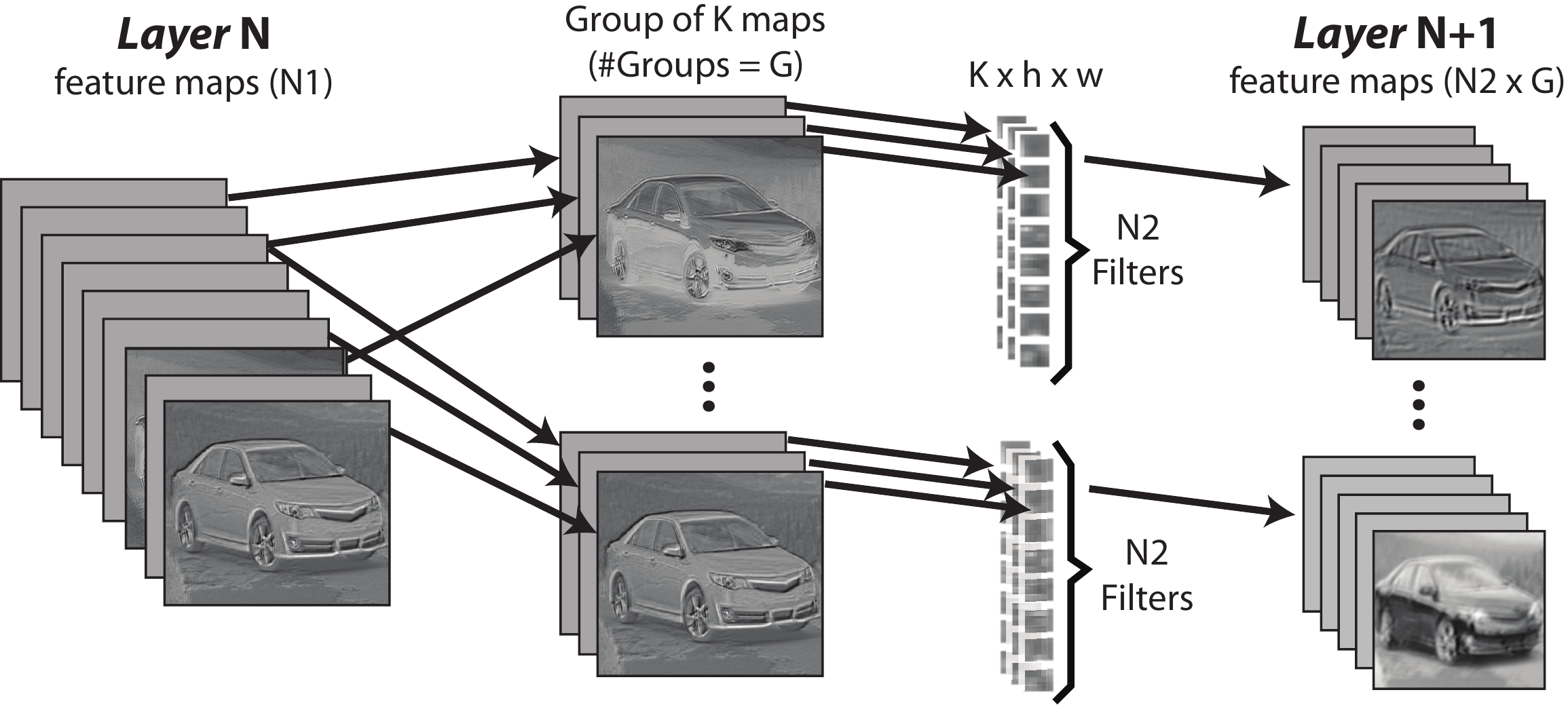}
\caption{Architecture of the connecting scheme from the first layer to the second layer. 
Different feature maps from the first layer are grouped together based on metrics (random, similarity, single, and full). 
Filters are learned from these groups individually with the clustering learning (CL) technique.
The filters are only applied to the group of feature maps that they are learned from.}
\label{fig-learnlayers}
\end{figure}

We investigate different techniques for grouping feature maps from layer $N$ into receptive fields (RF).
These receptive fields are the input of one or multiple (fanin or $K$) maps in layer $N+1$. In order to learn the filters for layer $N+1$ convolutional module, we perform clustering learning only with the receptive field (and not all $N1$ maps). We show a diagram in Fig. \ref{fig-learnlayers}. We learn $N2$ filters per group for a total of $N2 \cdot G$ maps in layer $N+1$.
We form these RF by four different methods, described below:

\begin{enumerate}

\item Grouping each maps individually (K = 1, learned RF). In other words, we create $N1$ groups which have only 1 feature map. 
$G = N1$ and $K = 1$ is used in this case (refer to Fig. \ref{fig-learnlayers}).

\item Grouping a small number of layer $N$ features that co-occur in one or multiple sets of feature maps from layer $N$ (K = 2, learned RF). 
In other words, we look for features that are highly activated in the same pixel, and we group these features into the RF.
In Fig. \ref{fig-learnlayers}, these are $G$ groups of $K$ maps.
The idea behind this step is to group features that occur often together in the data \cite{masquelier2007learning}.

\item Selecting a small number of layer $N$ features randomly (K = 2, random RF).
This test is very similar to the second test, but we didn't use any similarity metric to group similar features together.
Instead, we group features randomly.

\item Grouping all input maps into 1 group fully connected).
This method is basically a full connection from the first layer to the second layer. 
$G = 1$ and $K = 32$ are used in this case.

\end{enumerate}

\subsection{Computing feature similarity}

Similarities of features are calculated to learn the RF. 
The main driving force for this idea is a bio-inspired model of learning in the mammalian brain.
It is widely believed that one of the possible physical mechanism of learning in the brain is the Hebbian method \cite{masquelier2007learning}.
This learning method suggests that a group of pre-synaptic neurons is strongly connected to a post-synaptic neuron if the group
(or receptive field, RF) can predict the response of the post-synaptic neuron.
In other words, neurons in the RF have strong connections to a neuron in layer $N+1$ if it can predict their response.
The learned RF group features that are highly activated together in the same location.
This group of highly activated features will then propagate a high value to the same location in the relative map of layer $N+1$.
This implements the Hebbian rule.

The correlation of activation of features is computed by creating a similarity matrix between all feature maps in one or more sample images from the datasets. By computing the normalized cross-correlation for all pixels in each image and sorting their similarities in descending order, we can define which feature maps are more similar to each other. In order to compute meaningful statistics of co-activation of features, we used several hundred images from each dataset.

\section{Previous work}
\label{sec-priorart}

In the literature, we found two deep learning papers that are related. 
Paper \cite{coates2012learning} groups features on each layer by similarity, but does not explicitly use receptive fields between layers, 
rather they flatten the data at each layer and re-apply clustering algorithms. 
This approach is equivalent to forming a full-connection between layers, 
which is less efficient because each feature is averaged with a large number of other features, thus reducing the signal-to-noise ratio of important features.
The main difference between paper \cite{coates2012learning} and our work is that we focus on learning smaller networks,
with much fewer number of filters (32 vs. 4096) in the first and successive layers. 
The paper \cite{boureau_2011} presents the interesting idea of learning pooling strategies that include feature maps with similarity in local feature space. 
This paper uses simple clustering procedures to cluster data in order to improve pooling in the feature space. 
In this respect, this paper presents similar ideas to the first technique we test in this paper, but is not applied to deep neural networks.

Several papers in the neural physiology and computational neuroscience literature report similar ideas for learning the receptive fields of complex cells \cite{masquelier2007learning,spratling2005learning,wiskott2002slow,wallis1997invariant}. None of these paper reports results close to the state-of-the-art in publicly available datasets of natural images.

\section{Methods}
\label{sec-methods}

We used the Torch7 software for all our experiments \cite{collobert_torch7_2011}, since this software can reduce training and learning of deep networks
by 5-10 times compared to similar Matlab and Python tools.
In addition it features a "Spatial" mode that allows a trained network to be efficiently applied to any size image, for quick demonstrations and applications.

\subsection{Input data}

We tested and obtained results using the CIFAR-10 \cite{krizhevsky_learning_2009} and the Street View House Numbers (SVHN) \cite{netzer_reading_2011} datasets.
The SVHN dataset has a training size of 73,257 32$\times$32 images and test size of 26,032 32$\times$32 images.
The CIFAR-10 dataset has a training size of 20,000 32$\times$32 images and a test size of 10,000 32$\times$32 images.
Both datasets offer a 10 categories classification task on 32$\times$32 size images.
The train dataset was in all cases learned to 100\% accuracy, and training was then stopped. 

We did not use the YUV color space in CIFAR-10 and SVHN because they were reporting $\approx 2-4\%$ loss in accuracy.
This is contrary to what has been reported by others \cite{jarrett_what_2009}.
We kept the images of CIFAR-10 and SVHN in their original RGB format.
Input data was first normalized by subtracting out the mean and dividing by the standard deviation and then whitened.

In our experiment we fed all RGB planes to the deep network.
We also subsampled the RGB data and concatenated it with the output of the deep network into a final vector.
This final vector was then passed to a 2-layer MLP classifier.

\subsection{Network architecture}
\label{sec-net-arch}

We experimented by training an unsupervised deep neural network with 2 layers, not counting pooling and normalization operations.
The two layers were composed of a two-dimensional convolutional linear filtering stage, a spatial max pooling stage, and a thresholding stage. 
The filters of the first two layers are generated with unsupervised clustering algorithms, as explained below. 
Using the naming convention in \cite{lecun_convolutional_2010}, for each layer $l$, $x_i$ is an input feature map, $y_i$ is an output feature map.
The input of each layer is a 3D array with $n_l$ 2D feature maps of size $n_{l1} \times n_{l2}$.
Each component (pixel, neuron) is denoted $x_{ijk}$.
The output is also a 3D array, $y_i$ composed of $m_l$ feature maps of size $m_{l1} \times m_{l2}$.

\begin{figure}
\includegraphics[width=5in]{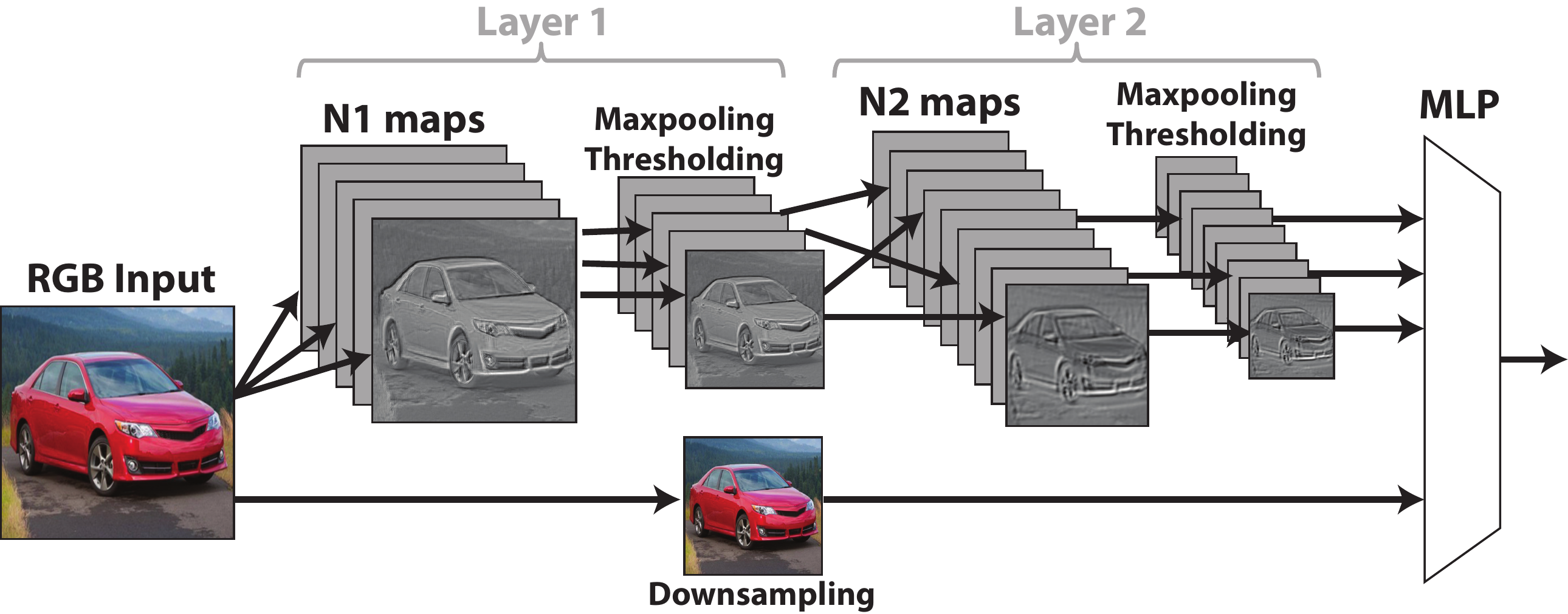}
\caption{Architecture of the CL network used in this paper: a 2 layer convolutional network with color bypass.}
\label{fig-netarch}
\end{figure}

The layers in the clustering learning network used the following sequence of operations:
\begin{enumerate}
\item Spatial Convolution module: performing convolutions on images with the learned CL filters: $y_{c_i}=\sum_i{k_{ij}\ast x_i}$,
where $\ast$ is the 2D discrete convolution operator.
\item Spatial Max Pooling module: $y_{p_i} = max_{n \times n}(y_{c_{ij}})$ with $n = 2$.
\item Threshold nonlinearity: $y_i = max(y_{p_i}, \theta)$ where the threshold $\theta = 0$.
\end{enumerate}

All networks used 32 filters on the first layer, and 512 filters on the second layer.
We use k-means clustering algorithm (we refer to this step as \textit{Clustering Learning algorithm})
to learn a set of 32 filters in the first layer, and 512 filters in the second layer.
The filter sizes on both layers were set to 5 $\times$ 5 pixels.
After the first convolution, 32 filters produced 32 feature maps each size 28 $\times$ 28.
Dimension of the feature maps decreased by 2 $\times$ 2 pooling with a stride of 2 pixels.
We used max pooling layer which is a bio-inspired approach to reducing the dimensionality of the data at each layer \cite{lampl2004intracellular}.
The output became 32 $\times$ 14 $\times$ 14 pixels.
In the second layer, 512 filters produced output of 512 $\times$ 10 $\times$ 10 feature maps which then max pooled again and the size became 512 $\times$ 5 $\times$ 5. 
We also subsampled the RGB data by 4 $\times$ 4 with a stride of 4 pixels which gave 3 $\times$ 8 $\times$ 8 pixels.
We then concatenated the subsampled RGB data with the output of the deep network into a final vector.
This final vector was then passed to a 2-layer MLP classifier.
The final classifier was fixed to 128 hidden units and 10 output classes for both CIFAR-10 and SVHN.




\section{Results}
\label{sec-results}

\begin {table}
\parbox{.5\linewidth}{
\caption{Results on CIFAR-10}
\label{tab:results-cifar10}
\begin{center}
\vspace*{-8pt}
\begin{tabular}{ll}
\multicolumn{1}{c}{\bf Architecture}  &\multicolumn{1}{c}{\bf Accuracy}
 \\ \hline \hline \\
1 layer (32 filters)			& $68.8\%$\\
2 layers (K = 1)				&$72.8\%$ \\
2 layers (K = 2, random RF)	&$73.2\%$ \\
2 layers (K = 2, learned RF)	&$71.2\%$\\
2 layers (Fully connected)		&$70.0\%$\\
\\ \hline  \hline
\end{tabular}
\vspace*{-15pt}
\end{center}
}
\parbox{.5\linewidth}{
\caption{Results on SVHN}
\label{tab:results-svhn}
\begin{center}
\vspace*{-8pt}
\begin{tabular}{ll}
\multicolumn{1}{c}{\bf Architecture}  &\multicolumn{1}{c}{\bf Accuracy}
\\ \hline \hline \\
1 layer (32 filters)			& $85.7\%$\\
2 layers (K = 1)                           &$87.6\%$ \\
2 layers (K = 2, random RF)	&$88.1\%$ \\
2 layers (K = 2, learned RF)	&$86.7\%$\\
2 layers (Fully connected)		&$86.4\%$\\
\\ \hline \hline
\end{tabular}
\vspace*{-15pt}
\end{center}
}
\end{table}

We presented results on CL for general-purpose vision system. We focused on the small number of filter to make our network scalable on hardware. We show results on CIFAR-10 and SVHN datasets with different connection techniques between the first and second layer.  We used the following parameters for each type of RF, as described in Section \ref{sec-main}:

K = 1, learned RF: we fixed the number of filters in the first layer to 32, $N1=32$. From each group, 16 filters are learned by CL, ($N2 = 16$). Therefore, in total we produce $N2 \cdot G = 32 \cdot 16 = 512$ filters. The size of filters is 1 $\times$ 5 $\times$ 5. 

K = 2, learned RF: we group 2 features together from the first layer. We again created 32 groups ($G = 32$ and $K = 2$), and from each group we learned 16 filters. Therefore, in total we again produce 512 filters but in this case the filters have size of 2 $\times$ 5 $\times$ 5. 

K = 2, random RF: all the numbers are kept the same as in the second test. With this test, we want to see if grouping the similar features gives any advantages over the random grouping.

Fully connected: in order to keep the number of filters same as the other tests we learned 512 filters from this 1 group which includes all the feature maps from the first layer. In this case, the size of filters is 32 $\times$ 5 $\times$ 5.

\begin{figure}
        \centering
         \label{fig-secondconnex}
        \begin{subfigure}[b]{0.3\textwidth}
                \centering
                \includegraphics[width=1.0in]{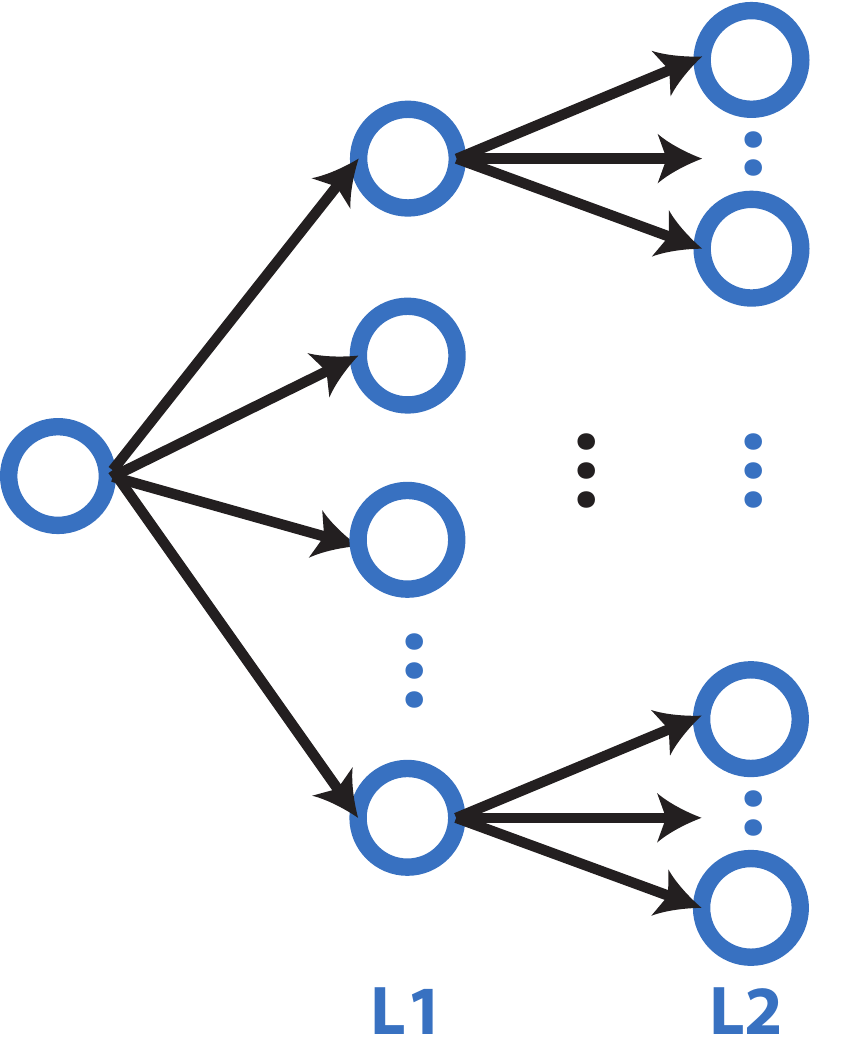}
                \caption{K = 1}
                \label{fig-secondconnex-fanin1}
        \end{subfigure}%
        ~
          \begin{subfigure}[b]{0.3\textwidth}
                \centering
                \includegraphics[width=1.0in]{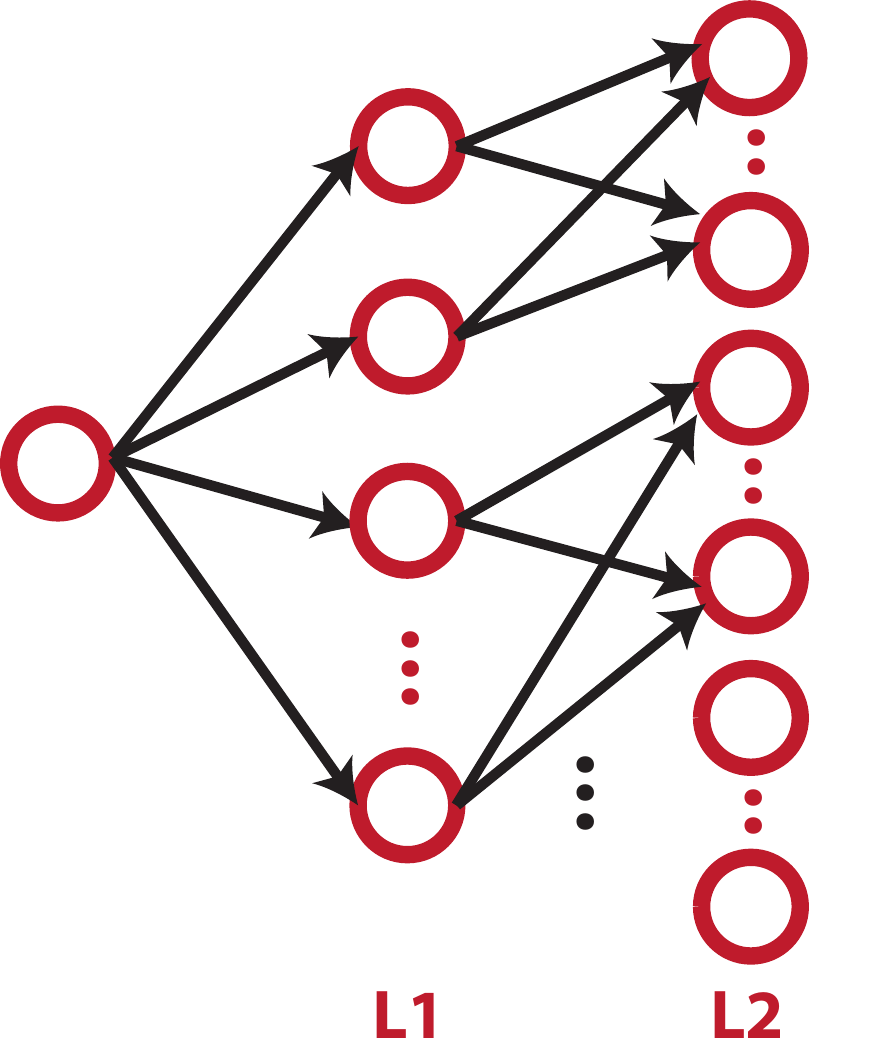}
                \caption{K = 2}
                \label{fig-secondconnex-fanin2}                
        \end{subfigure}%
        \begin{subfigure}[b]{0.3\textwidth}
                \centering
                \includegraphics[width=1.0in]{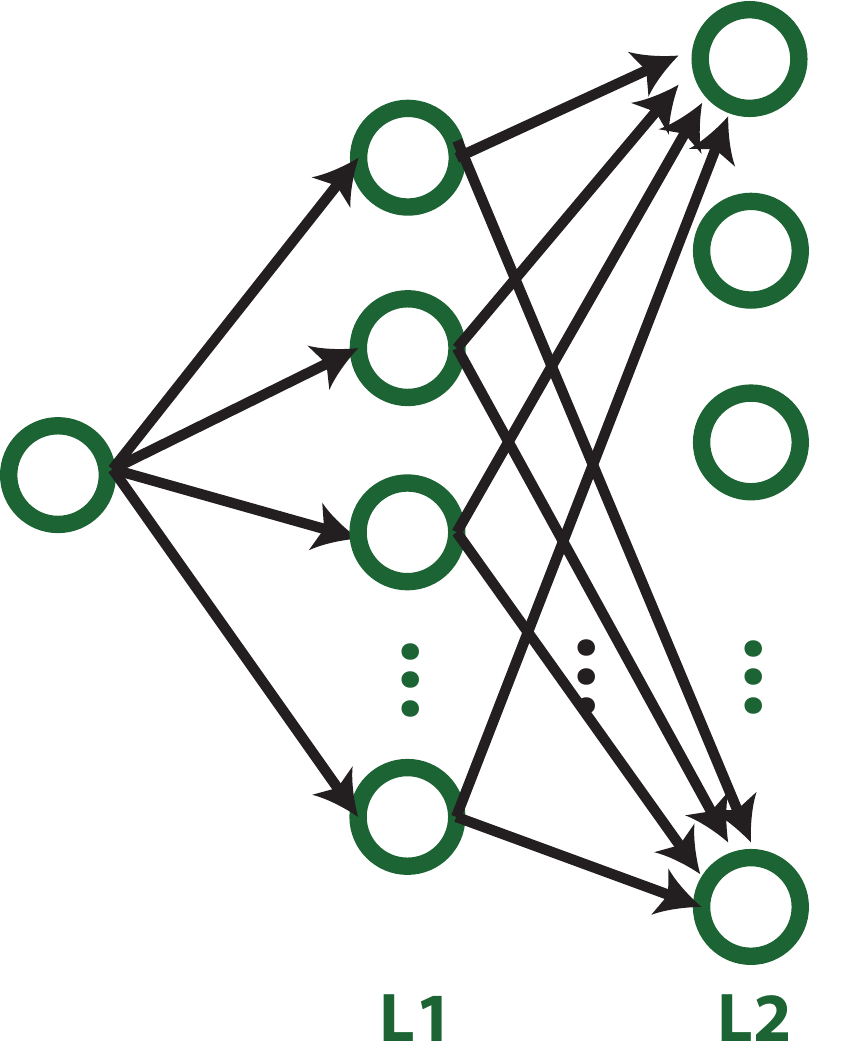}
                \caption{Fully connected}
                \label{fig-secondconnex-full}
        \end{subfigure}
        \caption{Simplified diagram of networks with different connection techniques.}
\end{figure}

As reported in the Table \ref{tab:results-cifar10} and \ref{tab:results-svhn}, 1 layer (32 filters) achieves $68.8\%$ accuracy on CIFAR-10 dataset and $85.7\%$ on SVHN dataset.
The accuracy on the CIFAR-10 is similar to the k-means algorithm with 200 filters from the paper \cite{coates_analysis_2011}.
We are obtaining the similar performance with only 32 filters because we use a different architecture in our network than \cite{coates_analysis_2011}.
We use 2 $\times$ 2 max pooling layer, whereas \cite{coates_analysis_2011} uses 7 $\times$ 7 subsampling pooling, which gives better results in a small network.
Another improvement comes from the downsampling of the image and feeding it to the classifier which increases the accuracy by $\approx 4\%$.
The similar bypass approach was used in a few other works \cite{hinton2012dropout,coates_selecting_2011}. 
This color bypass delivers strong color information to the final classifier.

The single RF architecture (K = 1, learned RF) can be interpreted as decision tree as shown in Fig. \ref{fig-secondconnex-fanin1}.
The decision tree architecture obtains a high accuracy of 72.8\% compared to 1 layer. This is due to the fact that the tree is able to refine the selectivity of each feature at each level.
This method handles only one feature from the previous layer instead of exploiting information from the group of features.

\begin{figure}
        \centering
        \label{fig-secondkernels}
        \begin{subfigure}[b]{0.5\textwidth}
                \centering
                \includegraphics[width=2.0in]{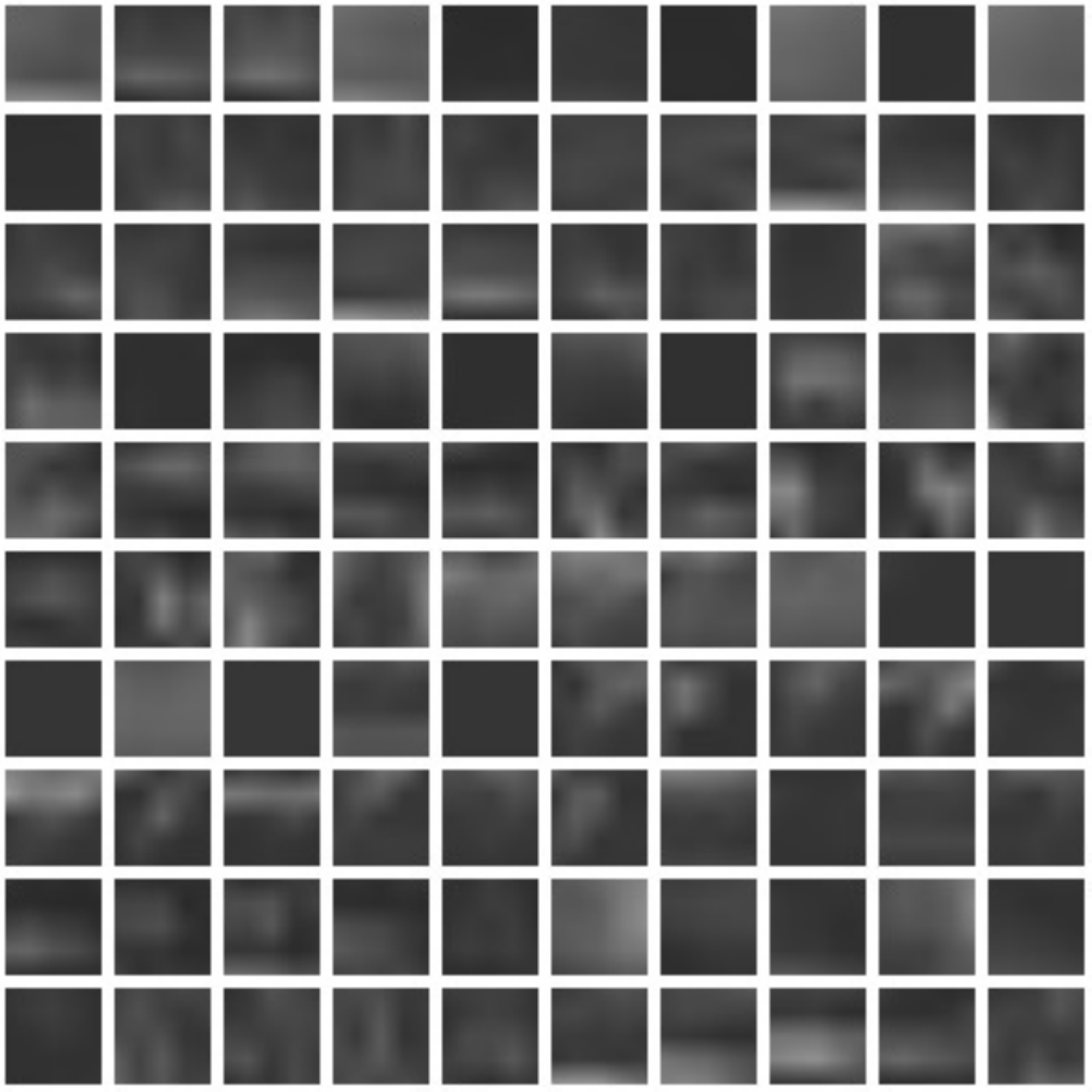}
                \caption{Filters learned with a full connection table}
                \label{fig-secondkernels-full}
        \end{subfigure}%
        ~
        \begin{subfigure}[b]{0.5\textwidth}
                \centering
                \includegraphics[width=2.0in]{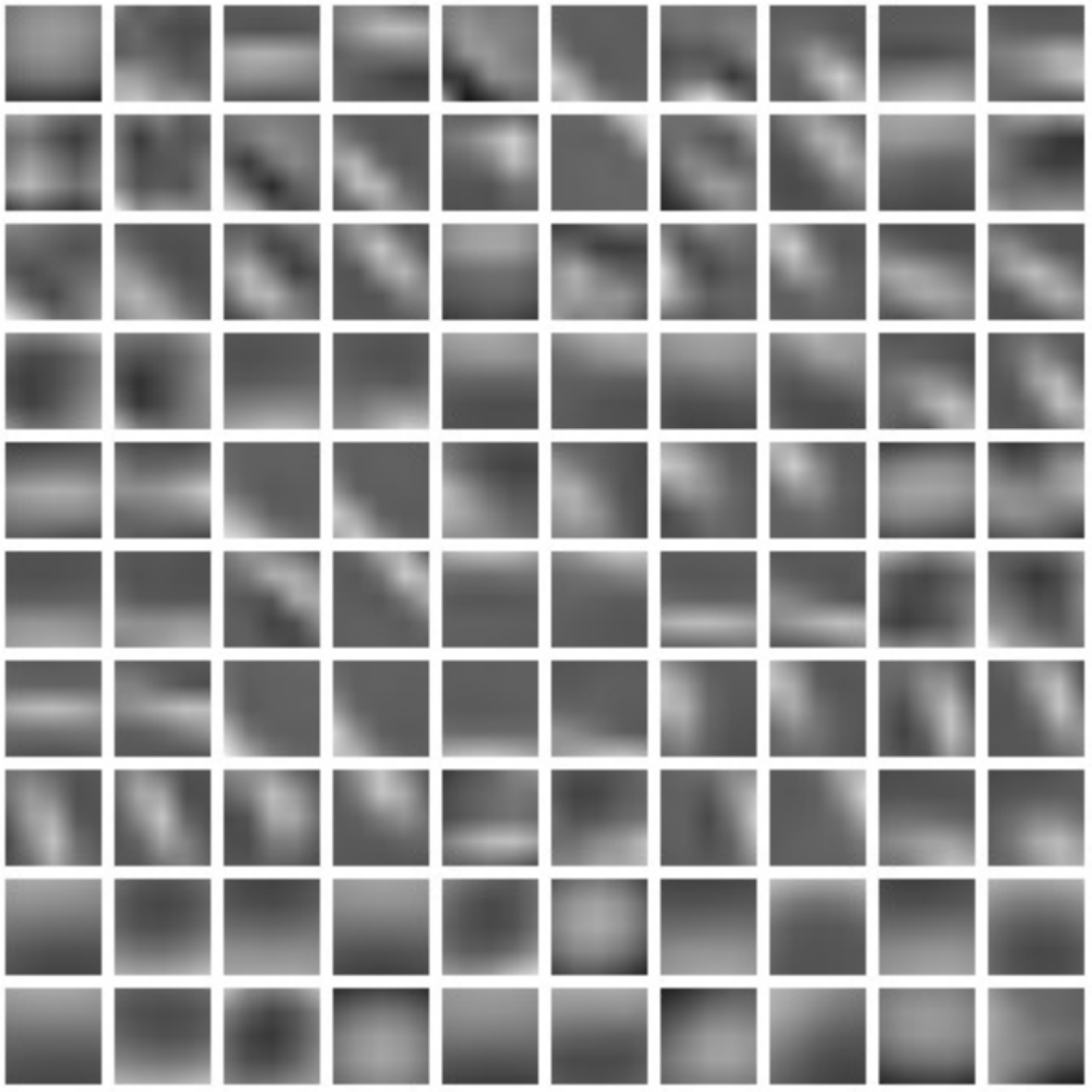}
                \caption{Filters learned with fanin 2 random connection}
                \label{fig-secondkernels-random}
        \end{subfigure}
        \caption{Second layer filters from CL for comparison.  The filters learned with the full connection table look uniform, while the filters learned with the K = 2 random connections table provide more specificity.}
\end{figure}

The RF with 2 feature maps (K = 2, learned RF) are compared to random combination (K = 2, random RF) to analyze the possible advantages of exploring different features in the same RF. Learned RF's are obtained  based on the similar ideas for learning the receptive fields of complex cells that are reported in  several papers in the neural physiology and computational neuroscience literature \cite{masquelier2007learning,spratling2005learning,wiskott2002slow,wallis1997invariant}.
Unfortunately in our tests we did not see an advantage in learning the RF compare to random groups of features for each RF.

\setlength{\tabcolsep}{10pt}

\begin{wraptable}{r}{0.5\columnwidth}
\vspace*{-15pt}
\caption{Results on CIFAR-10, 2 layers with different number of fanin.}
\vspace*{-5pt}
\begin{centering}
\begin{tabular}{ll}
\multicolumn{1}{c}{\bf Number of connections}  &\multicolumn{1}{c}{\bf Accuracy}\\
\hline \hline \\
K = 1		&$72.8\%$\\
K = 2		&$73.2\%$\\
K = 4		&$71.1\%$\\
K = 8		&$71.1\%$\\
K = 16		&$70.0\%$
\\ \hline \hline 
\label{tab:fanin}
\end{tabular}
\vspace*{-25pt}
\end{centering}
\end{wraptable}

The fully connected RF use filters obtained by CL across all 32 input feature maps. This architecture suffers from the curse of dimensionality because of the large 32 $\times$ 5 $\times$ 5 high-dimensional feature space, and clustering learning is not able to find selective RF and groups.
Fig. \ref{fig-secondkernels-full} supports this argument because the filters reported do not have distinct selectivity, but are rather plainly uniform. A similar argument was also made in the paper \cite{coates2012learning}.

Since the random grouping of the features gave the best results on CIFAR-10 and SVHN as shown in the Tables \ref{tab:results-cifar10} and \ref{tab:results-svhn}, we ran experiments to measure the effect of varying the fanin of the connection table.
The results are presented in the Table \ref{tab:fanin}. A fanin of 2 provides higher accuracy than a one-to-one connection table, but increasing the fanin above 2 leads to a loss in accuracy.

The current state of the art on CIFAR-10, 88.79\% accuracy, was reported in the paper \cite{dan2012multicolnet} with a 10-layer multi-column deep network. However,
such a massive network is not suitable for real-time applications. Our work focuses on small networks that can run in real-time.

We believe that random grouping provides powerful results in the case of small number of input filters. In our work we used only 32 filters in order to perform real-time execution on standard hardware. If the number of filters is higher, as in  \cite{coates2012learning} and similar work, then grouping the RF by similarly is effective. In the future we plan to use a larger number of filters organized topographically, and pool features not only spatially, but also in the feature dimension. The topographic organization will allow to use a 3D pooling stage, and we will then re-test learning of RF in this case.

\section{Conclusion}
\label{sec-conc}

We have presented an analysis of four different techniques for selecting receptive fields between layers of deep neural networks.
These techniques consisted of one-to-one connection, two-to-one connections and full-connected networks.
For two-to-one connections, we used random grouping as well as grouping based on the similarity of the feature maps from the first layer.  We tested these networks on the CIFAR and SVHN datasets and showed the fully-connected network performs poorly with unsupervised learning. We also showed that learning the connections based on the co-occurrences of  the features does not confer an advantage over a random connection in small networks.
We expect this work to be helpful to improve the efficiency of connections between the layers of deep small neural networks for the further study.

\subsubsection*{Acknowledgments}
We are especially grateful to the the Torch7 developing team, in particular Ronan Collobert, who first developed this great, easy and efficient tool,
Clement Farabet, Koray Kavukcuoglu, Leon Bottou.
We could not have done any of this work without standing on these giants shoulders.
We also thank Soumith Chintala for his help with Torch and the nice discussions.

\bibliography{clustering}
\bibliographystyle{unsrt}

\end{document}